\documentclass[10pt,twocolumn,letterpaper]{article}

\usepackage[pagenumbers]{cvpr} % To force page numbers, e.g. for an arXiv version
\usepackage{graphicx}
\usepackage{amsmath}
\usepackage{amssymb}
\usepackage{booktabs}

\usepackage{comment}
\usepackage{nicefrac}
\usepackage{eufrak}
\usepackage{float}
\usepackage[pagebackref,breaklinks,colorlinks]{hyperref}
\newcommand{\name}{\text{AlteredAvatar}}

\usepackage[noend]{algorithmic}
\usepackage[ruled]{algorithm2e} % For algorithms
\usepackage{ulem} 

\SetAlFnt{\small}
\SetAlCapFnt{\small}
\SetAlCapNameFnt{\small}
\SetAlCapHSkip{0pt}

\DeclareUnicodeCharacter{202A}{!!!FIX ME!!!} 
\begin{document}
% Title portion
\title{$\name$: Stylizing Dynamic 3D Avatars with Fast Style Adaptation}

\author{
  Thu Nguyen-Phuoc
  \qquad\qquad Gabriel Schwartz
  \qquad\qquad Yuting Ye \\
  Stephen Lombardi
  \qquad\qquad Lei Xiao \\
  Reality Labs Research, Meta \\
}
\twocolumn[{%
\renewcommand\twocolumn[1][]{#1}%
\maketitle
\vspace{-1cm}
\begin{center}
    \centering 
    \includegraphics[width=\textwidth]{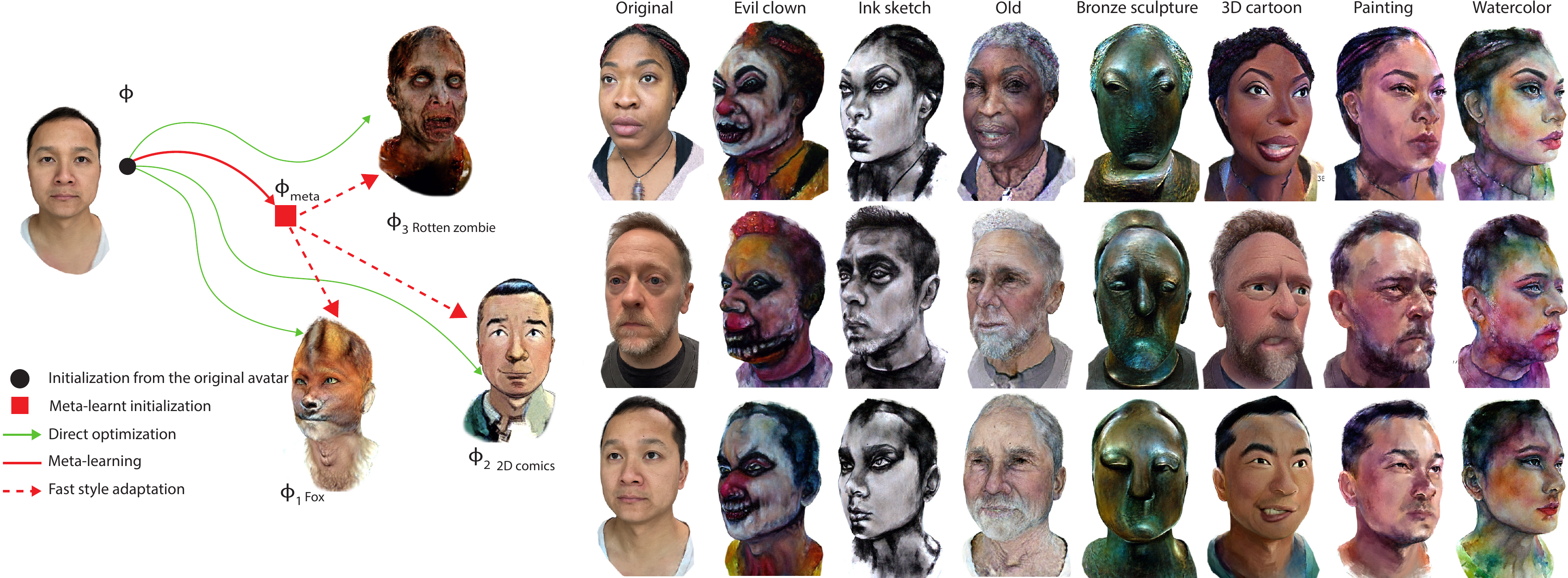}
    \vspace{0cm}
    \captionof{figure}{\label{fig:teaser} Given learnt photorealistic dynamic 3D avatars, AlteredAvatar stylizes the avatars to match a reference style image or a textual description. AlteredAvatar can fast adapt to novel styles in a small number of update steps, and generates consistent stylized results across different viewpoints and expressions.}
    \vspace{-0.cm}
\end{center}
}]
\begin{abstract}
This paper presents a method that can quickly adapt dynamic 3D avatars to arbitrary text descriptions of novel styles. Among existing approaches for avatar stylization, direct optimization methods can produce excellent results for arbitrary styles but they are unpleasantly slow. Furthermore, they require redoing the optimization process from scratch for every new input. Fast approximation methods using feed-forward networks trained on a large dataset of style images can generate results for new inputs quickly, but tend not to generalize well to novel styles and fall short in quality. We therefore investigate a new approach, AlteredAvatar, that combines those two approaches using the meta-learning framework. In the inner loop, the model learns to optimize to match a single target style well; while in the outer loop, the model learns to stylize efficiently across many styles. After training, AlteredAvatar learns an initialization that can quickly adapt within a small number of update steps to a novel style, which can be given using a text description, a reference image, or a combination of both. We show that AlteredAvatar can achieve a good balance between speed, flexibility and quality, while maintaining consistency across a wide range of novel views and facial expressions. 
\end{abstract}

\maketitle

\section{Introduction}
%Why avatar
Creating high-quality human face models has a wide range of applications in the movie and film industry.
Recently, social telepresence applications in virtual reality (VR) and mixed reality (MR) have opened up new opportunities for highly-accurate and authentic avatars that can be driven from users' expression input.
While photorealistic avatars are required in certain applications, stylized avatars would be desired in other scenarios.
Imagine being able to change one's avatar as one moves between different events and applications in VR: one can put on Goth makeup to attend a virtual concert, turn into an alien to attend a virtual Comic Con, or turn into a zombie to play VR zombie games.

In this paper, we focus on the problem of stylizing dynamic 3D head avatars for VR/MR applications. 
This presents a unique requirement for the choice of avatar representation: the avatars must faithfully capture the identity and expression of the users driving them, can be rendered from a wide range of novel views, and be efficient to render in high fidelity.
Most work in 3D avatar stylization relies on static mesh or tri-planes representation \cite{3dportrait,datid3d,abdal20233davatargan}. Other work adopts neural radiance fields which has high quality but slow to render \cite{nerfart, avatarclip, instructnerf2023}.
Therefore, we choose to work with Instant Avatar \cite{ICA}, a learnt 3D dynamic avatar representation based on U-Nets \cite{unet}.
% Compare to Neural Radiance Fields (NeRFs) and their dynamic variations
Instant Avatars can be efficiently rendered at high fidelity, and more importantly, faithfully capture the eye gaze and dynamicexpressions of the original identity \cite{mvp, ICA}, making them suitable for downstream MR/VR applications.

Until very recently, the majority of neural style transfer work can only work with style guidance from a single or a few 2D image \cite{neuralStyleTransfer, Li_2016_CVPR, yang2022Vtoonify, dualstylegan, men2022dct}.
With \name, we want to enable users to intuitively describe their desired style using texts, in addition to images.
This greatly opens up the possibilities of target styles, ranging from abstract style concepts such as "Impressionist movement" to very concrete ones such as a particular style of a famous painting or movie.
Therefore, we propose to use CLIP \cite{CLIP}, a model pre-trained for multi-modal language and vision reasoning with contrastive learning.
Additionally, CLIP has shown the ability to capture semantic concepts of visual depictions well \cite{vinker2022clipasso, Jain_2021_ICCV}. This produces stylized results that are semantically consistent with the input avatar and the target style textual description, unlike previous image-based stylization approaches that require both the content and style input images to be semantically similar for the best quality.

Work on neural stylization can be divided into two approaches: direct optimization, and training stylization networks. 
Direct optimization produces high quality results, but can be time and computationally intensive.
More importantly, for every new style or new content image, one has to do the optimization from scratch, making them impractical for product deployments. 
Stylization networks aim to solve this issue by training a network on large datasets of content and style images using perception loss \cite{Johnson2016Perceptual, huang2017adain} or GAN loss \cite{yang2022Vtoonify,dualstylegan}. 
At test time, new styles and contents can be used to synthesize the results without any retraining, but usually at the cost of degraded visual quality of the stylized results.

We propose $\name$, a third approach that aims to strike a balance between the computational time, the flexibility to adapt to novel styles, and the quality of the stylized results.
In particular, we aim to learn an avatar representation that can be quickly adapted to unseen styles after a small number of update steps.
It is worth noting that our approach does not add any extra parameters to the avatar representations.
Instead, using a meta-learning framework, we only fine-tune a small subset of the avatar network parameters to learn a good weight initialization that can rapidly adapt to new styles.
Previously, meta-learning has been used to learn representations that can quickly adapt to novel material \cite{metapperance}, 3D static scenes \cite{bergman2021metanlr, tancik2020meta}, talking head videos \cite{zhang2022metaportrait} or styles for image neural style transfer \cite{metastyle}.
This is different from approaches that are based on hypernetworks \cite{ha2017hypernetworks}, which train a separate network to predict a new set of network weights for every new input.
Hypernetwork-based approaches can be memory-intensive and tend to produce less competitive results \cite{8100228, chiang2021stylizing, hyperdomainnet}.

In summary, our proposed $\name$ offers the first dynamic 3D avatar stylization framework that can quickly adapt to match novel styles given as text descriptions or a combination of both texts and images.
The contributions of $\name$ are threefold:
\begin{itemize}
\item We introduce a meta-learning approach for stylizing dynamic 3D avatars, enabling fast adaptation to any new style using only a small number of update steps.
\item We show that using CLIP features in $\name$ provides an expressive and flexible framework for semantically-meaningful stylization using a textual description, or a combination of both reference style text and image.
\item Through extensive experiments, we demonstrate that $\name$ produces stylized dynamic 3D avatars in high resolution with consistency across views and expressions, with the flexibility to quickly adapt to novel styles.
\end{itemize}

\section{Related work}
\subsection{Neural stylization}
Neural style transfer methods, including portrait stylization methods, can be divided into two categories: direct optimization methods and learning-based methods.
Optimization methods \cite{neuralStyleTransfer,Li_2016_CVPR, stylemesh, SNeRF,3dportrait,huang_2021_3d_scene_stylization} generate impressive results that match the content and styles images well, but are slow due to their iterative nature. 
Moreover, they lack flexibility since the results are overfitted to a particular style or content, and for every new input, one has to do the optimization from scratch.

Learning based methods \cite{Johnson2016Perceptual, WCT-NIPS-2017,li2018learning,chiang2021stylizing,dualstylegan, yang2022Vtoonify} instead train a style-transfer network on a style image dateset and can quickly generate results for new inputs without recomputation, even in real time \cite{huang2017adain}.
However, their results only work well for the styles that are used to train the networks \cite{Johnson2016Perceptual, instanceNorm, dualstylegan, yang2022Vtoonify} or have worse visual quality \cite{huang2017adain,chiang2021stylizing}.
Recent work adopts meta-learning to learn a model that can both adapt quickly to new arbitrary inputs and generate results with high visual quality \cite{metastyle, 8578939}.
Their results are promising but at the cost of increased network parameters to train \cite{8578939}, and only work with single 2D images \cite{metastyle, 8578939}.

With $\name$, we aim to learn dynamic 3D avatar representations that can quickly adapt to arbitrary styles using a single text description or image, while maintaining the style consistency across views and expressions. 
This can be seen as a third approach to stylization that strikes a balance between speed, flexibility to novel styles, and the quality of the results.
%%%%%%%%%%%%%%%%%%%%%%%%%%%%%%%%%%%%
%%%%%%%%%%%%%%%%%%%%%%%%%%%%%%%%%%%%
%%%%%%%%%%%%%%%%%%%%%%%%%%%%%%%%%%%%
%%%%%%%%%%%%%%%%%%%%%%%%%%%%%%%%%%%%
%%%%%%%%%%%%%%%%%%%%%%%%%%%%%%%%%%%%
%%%%%%%%%%%%%%%%%%%%%%%%%%%%%%%%%%%%
\subsection{Text-guided synthesis}
With recent advancements in text-to-image models \cite{rombach2021highresolution, dalle, imagen} and pre-trained language-vision models \cite{CLIP}, there has been an explosion of work in text-to-image, text-to-video, and text-to-3D synthesis \cite{rodin, dreamfields, avatarclip, clipmesh,clipvox,styleclip, text2mesh, dreamfusion, lin2023magic3d, rodin}.
Additionally, work in domain adaptation or image editing combines a pre-trained StyleGAN \cite{Karras_2019_CVPR} and CLIP to transfer photorealistic images to new domains using only texts without any (or only one) image training data \cite{stylegannada, hyperdomainnet, alaluf2021hyperstyle, datid3d, targetclip, styleclip, zhu2021mind}.
For stylization, using text descriptions provides users with a very intuitive interface to express their desired styles.
Moreover, textual descriptions can be very expressive and cover a wide range of styles.
This can be a specific art movement such as \textit{"Cubism"} or \textit{"Impressionism"}, or more descriptions such as \textit{"vintage poster art using only primary colours"}.

Recent work in text-guided 3D avatar stylization \cite{nerfart, aneja2022clipface,instructnerf2023, datid3d, zhang2023dreamface} shows the rapid advancement in text-guided avatar stylization.
However, ClipFace \cite{aneja2022clipface} only focuses on stylizing the texture of the avatar, although geometry has been acknowledged to be an important factor of style \cite{Kim20DST, 9577906}.
Meanwhile, RODIN \cite{rodin} can only generate static avatars with a particular aesthetics due to its reliance on a synthetic training dataset.
Similarly, DATID-3D \cite{datid3d} can adapt 3D faces to a novel style using a text-to-image diffusion-guided generative face model, but the faces are limited only to more frontal views and have fixed expressions.
Finally, DreamFace \cite{zhang2023dreamface}, despite its impressive text-to-face results, currently does not support personalizing to individual users's identities.
Meanwhile, concurrent work, NeRF-Art \cite{nerfart} and Instruct-NeRF2NeRF \cite{instructnerf2023}, only shows results with static avatars. Moreover, they are memory-intensive and slow to render due to their reliance on NeRF, and still rely on a lengthy optimization process for every new input.

Here, $\name$ works with 3D dynamic avatars that can be driven by users and rendered from a wide range of novel views. 
In addition to text, we show that interesting and novel styles can come from a combination of image and text style descriptions.
Finally, $\name$ focuses on adapting to novel styles using a small number of steps.
%%%%%%%%%%%%%%%%%%%%%%%%%%%%%%%%%%%%
%%%%%%%%%%%%%%%%%%%%%%%%%%%%%%%%%%%%
%%%%%%%%%%%%%%%%%%%%%%%%%%%%%%%%%%%%
%%%%%%%%%%%%%%%%%%%%%%%%%%%%%%%%%%%%
%%%%%%%%%%%%%%%%%%%%%%%%%%%%%%%%%%%%
%%%%%%%%%%%%%%%%%%%%%%%%%%%%%%%%%%%%
%
\subsection{Meta-Learning in computer graphics}
Inspired by the human ability to quickly adapt to new skills by drawing from past experience and knowledge, Meta-Learning is a group of methods that focuses on learning to learn. 
Meta-Learning algorithms typically take in a distribution of tasks, where each task is a learning problem, and produce a quick learner — a learner that can quickly adapt to new tasks from a small number of examples.
Note that this is different from work that use hypernetworks \cite{ha2017hypernetworks} for generalisation \cite{alaluf2021hyperstyle, hyperdomainnet}, which learns to directly predict a new set of weights for every new input, rather than learning a good initialization for fast adaptation.
Being a relatively new area of research in deep learning, applications of Meta-Learning in computer graphics are less extensive than other types of learning algorithms.
MetaNLR++ \cite{bergman2021metanlr} and Tancik {\emph{et al.}} \cite{tancik2020meta} use meta-learning to learn shape priors for fast training of neural lumigraph representations and coordinate-Based neural representations respectively.
Metappearance \cite{metapperance} applies meta-learning to a wide range of visual appearance reproduction problems such as textures, Bidirectional Reflectance Distribution Functions (BRDFs), or the entire light transport of a scene. 
More recently, MetaPortrait \cite{zhang2022metaportrait} offers fast personalized talking-head generation by training their networks under the meta-learning framework.

The most related work to ours is MetaStyle \cite{metastyle} that adopts meta-learning for 2D image style transfer, achieving a good trade-off among speed, flexibility, and quality for arbitrary artistic style transfer.
In this work, we apply meta-learning on the problem of 3D avatar stylization.
To the best of our knowledge, $\name$ is the first approach on dynamic 3D avatar stylization that focuses on fast adaptation to novel styles, generating results in a few update steps that are comparable with those that perform stylization optimization from scratch.

\section{Method}
\begin{figure}
\centering
	\includegraphics[width=\linewidth]{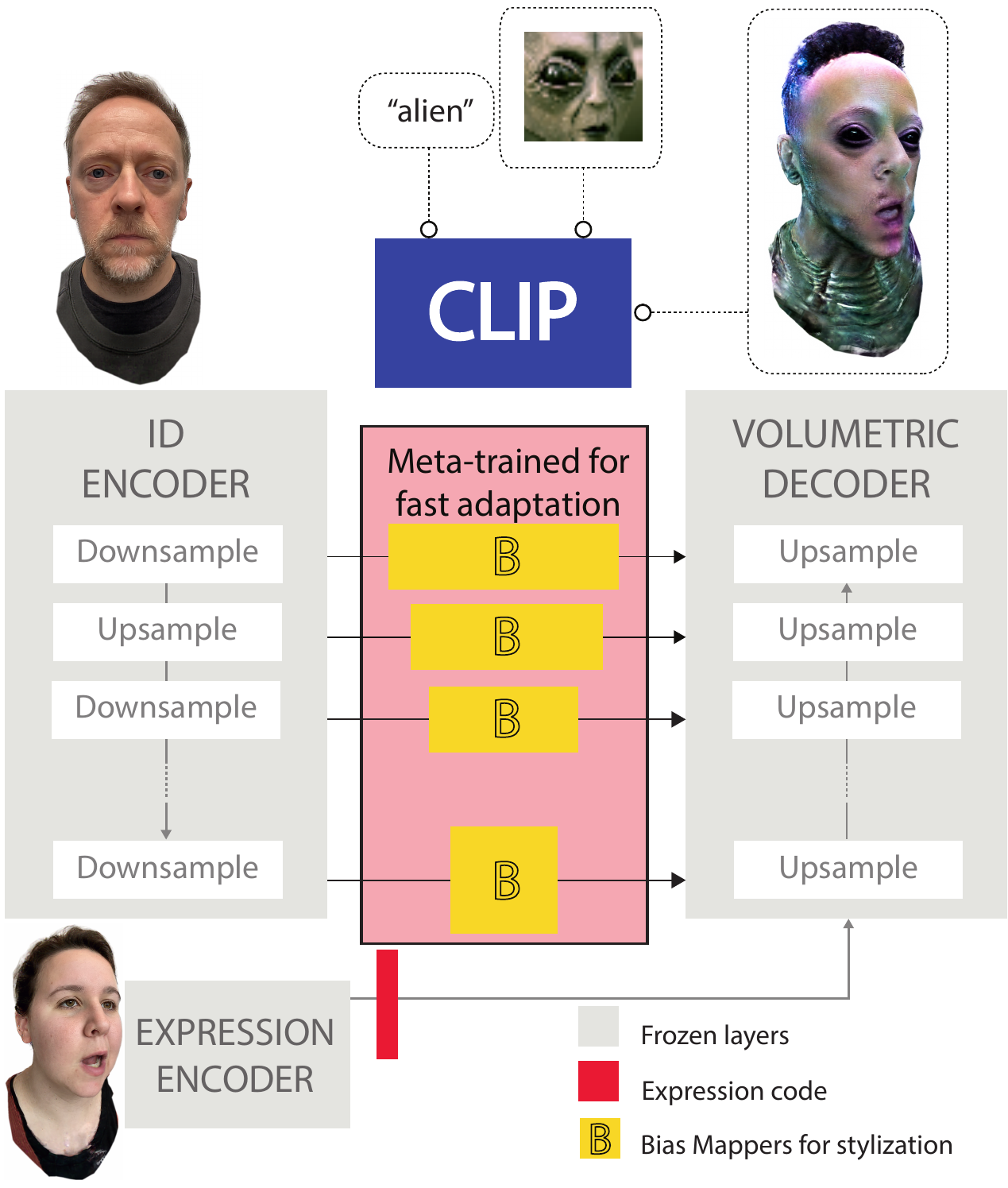}
	\caption{\label{fig:network}
		\textbf{Overall pipeline for $\name$}. Taking the Instant Avatar~\cite{ICA} as the avatar representation, we only use the weights of its bias mappers (yellow blocks) during the stylization process, and keep the rest of the network (the remains of the Identity Encoder, Expression Encoder, and Volumetric Decoder) frozen. Rendered images of the stylized avatar (from different views and expressions) are given to a pre-trained CLIP model to compute the stylization loss $\mathcal{L}_\text{style}$ against an input style text description or a target style image. The stylized avatar can be driven using expressions from the same or from another avatar.}
\end{figure}
Given a pre-trained photorealisitc dynamic 3D avatar $\mathcal{M}$, we aim to manipulate its appearance and geometry such that rendered images of the avatars match the style from a reference image $\textbf{I}_{style}$ or a text description $\mathbf{t}_{style}$.
Additionally, rendered images of the stylized avatar $\hat{\mathcal{M}}$ from different views and expressions should be consistent. 
In this work, we adopt the Instant Avatar architecture \cite{ICA} for its rendering efficiency and ability to represent high-quality dynamic avatars.
We propose to use pre-trained CLIP models \cite{CLIP} as our source of style supervision, instead of VGG \cite{VGG} like other neural stylization work.
Finally, we propose a meta-learning approach to learn an avatar representation that can quickly adapt to an unseen style using a small number of update steps, while achieving the high fidelity and maintaining the avatar's ability to be animated to different expressions (see Figure \ref{fig:teaser}-Left).

\subsection{Preliminaries}
\subsubsection{Dynamic 3D avatar representation}\label{sec:ICA}
Without loss of generalization, here, we use the Instant Avatar representations \cite{ICA} for avatar stylization. 
It is a lightweight version of Mixture of Volumetric Primitives (MVP)~\cite{mvp}, a hybrid representation that combines the expressiveness of volumetric representations with the efficiency of primitive-based rendering for high-quality dynamic 3D contents.
Instant Avatar can represent dynamic photorealistic avatars with very high fidelity, and explicitly disentangle an avatar's identity from its expression.
Instant Avatar is essentially a U-net \cite{unet} that maps 2D conditioning data to 3D volumetric slabs, which are then ray-marched to generate images of the avatar.
The network is trained using an image-based loss with posed image data, and is fine-tuned to each individual to achieve the highest level of authenticity.
Compared to pure NeRF-based models such as NerFACE \cite{Gafni_2021_CVPR}, Instant Avatar needs less data, significantly less time and memory to train, and less time to render.

Notably, Instant Avatar contains two encoders: an Identity Encoder and Expression Encoder.
The Identity Encoder maps 2D position maps (representing a mesh) and unrolled 2D texture maps (representing appearance) of a specific person to multi-scale bias maps using blocks of 2D-convolutional bias mappers.
These maps are decoded to a modified base-mesh and volumetric slabs, and rendered to 2D images using ray marching.
In summary, given a trained Instant Avatar of a specific person $\mathcal{M}_{\phi}$ and a deterministic ray marcher $R$, we can describe its rendered image $\mathbf{I}_\theta$ from a particular view $\theta$, given identity bias maps $\mathbf{x}_i$ (from the Identity Encoder) and expression code $\mathbf{z}_e$ (from the Expression Encoder) as:
\begin{equation}\label{eq:avatars}
    \mathbf{I}^{\theta}= R\left(\mathcal{M}_{\phi}(\mathbf{x}_\mathrm{i}, \mathbf{z}_\mathrm{e});\theta\right).
\end{equation}
Since we are interested in changing the shape and appearance of an avatar to match a desired style, we only manipulate the weights of the Identity Encoder during the stylization process, and keep the Decoder and Expression Encoder frozen (see Figure \ref{fig:network}).
Additionally, we empirically found that using only the blocks of bias mappers within the Identity Encoder lead to the same amount of stylization effects compared to using all of the layers.
Therefore, we opt to use them instead.
From now on, when we use $\mathcal{M}_{\phi}$ to describe an avatar network, we only focus on the weights of the bias mappers in the Identity Encoder. 
\subsection{CLIP-guided avatar stylization}
For avatar stylization, we want to find an avatar $\hat{\mathcal{M}_{\phi}}$ that matches a target text description $\textbf{t}_{tgt}$ or style image $\textbf{I}_\text{style}$, while maintaining part of its original identity.
Without any 3D style data, we leverage multi-view rendered images of a stylized avatar $\hat{\mathcal{M}_{\phi}}$ to compute the stylization loss.
Unlike other neural style transfer work that uses VGG features \cite{VGG}, $\name$ uses CLIP features \cite{CLIP} to compute the stylization loss. 
Given a rendered image $\mathbf{I}_{tgt}$ of $\hat{\mathcal{M}_{\phi}}$ from a random view and with a random expression, and a target style description $\mathbf{t}_{tgt}$, a simple stylization loss can be written as the cosine similarity $(\langle \cdot , \cdot \rangle)$ between the CLIP's embeddings of the avatar's rendered image and the text description:
\begin{equation}
\begin{aligned}
\label{eq:global}
\mathcal{L}_\text{CLIP}=\sum_{\mathbf{I}_{tgt}}\left[1 - \langle\, \mathcal{E}_{i}({\mathbf{I}_{tgt}}), \mathcal{E}_{t}(\mathbf{t}_{tgt})\rangle\right]
\end{aligned}
\end{equation}
where $\mathcal{E}_i$ and $\mathcal{E}_t$ are the pre-trained CLIP image and text encoder, respectively. 
\subsubsection{Identity loss}
For avatar stylization, we want to meaningfully change the geometry and the appearance of the avatar to match a desired style while maintaining the identity of the original avatar.
Therefore, we adopt the identity loss $\mathcal{L}_\text{ID}$ to control the amount of identity preservation in the stylized results.
This loss essentially compares the structures, represented as a self-similarity matrix, of the stylized image $\textbf{I}_{tgt}$ with that of the original RGB image $\textbf{I}_{src}$.
Specifically, for each image, we extract its CLIP's $N$ spatial tokens from the deepest layer, and compute a self-similarity matrix $\mathbf{S}\left(\mathbf{I}\right) \in \mathbb{R}^{N \times N}$. Each matrix element $\mathbf{S}\left(\mathbf{I}\right)_{i,j}$ is defined by:
\begin{equation}
\begin{aligned}
\mathbf{S}\left(\mathbf{I}\right)_{i,j} = 1 - \langle\, n_{i}(\mathbf{I}), n_{j}(\mathbf{I})\rangle.
\end{aligned}
\end{equation}
where $n_i \in \mathbb{R}^{768}$ is the $i^th$ token of an image $\mathbf{I}$.
Finally, the identity loss $\mathcal{L}_\text{ID}$ is defined as the Frobenius norm between the self-similarity matrix of $\textbf{I}_{tgt}$ and $\mathbf{I}_{src}$:
\begin{equation}
\begin{aligned}
\mathcal{L}_\text{ID}= \| \mathcal{S}(\mathbf{I}_{tgt}) - \mathcal{S}(\mathbf{I}_{src})\|_F.
\end{aligned}
\end{equation}
In summary, the final stylization loss is given by:
\begin{equation}\label{eq:loss_sty}
\begin{aligned}
    \mathcal{L}_{\text{style}} = \lambda_{\text{CLIP}}\mathcal{L}_\text{CLIP} + \lambda_{\text{ID}}\mathcal{L}_{\text{ID}}
\end{aligned}
\end{equation}
%%%%%%%%%%%%%%%%%%%%%%%%%%%%%%%%%%%%
%%%%%%%%%%%%%%%%%%%%%%%%%%%%%%%%%%%%
%%%%%%%%%%%%%%%%%%%%%%%%%%%%%%%%%%%%
%%%%%%%%%%%%%%%%%%%%%%%%%%%%%%%%%%%%
%%%%%%%%%%%%%%%%%%%%%%%%%%%%%%%%%%%%
%%%%%%%%%%%%%%%%%%%%%%%%%%%%%%%%%%%%
\subsection {Fast style adaptation}
To stylize an avatar $\mathcal{M}_{\phi}$ with weights $\phi$ to a particular style $s$, we minimize the stylization loss $\mathcal{L}_\text{style}$ to find the weights $\hat{\phi}_s$ of the stylized avatar:
\begin{equation}
\label{eq:loss_sty_meta}
\begin{aligned}
    \hat{\phi}_s = \underset{\phi_s}{\text{argmin}} \mathcal{L}_{\text{style}}(\mathcal{M}_{\phi_s}, s)
\end{aligned}
\end{equation}
Usually this is done by performing gradient descent with a gradient-based optimization method. In this case, we use stochastic gradient descent (SGD) for $K$ steps from the initialization $\phi$ to find $\hat{\phi}_{s}$, which can be formulated as:
\begin{equation}
\label{eq:SGD}
\begin{aligned}
    \hat{{\phi}}_s^K = \text{SGD}(\mathcal{L}_\text{style}(\mathcal{M}_{\phi_s}, s), K)
\end{aligned}
\end{equation}
For every new style $s$, the stylization optimization needs to be done from scratch to find new weights $\hat{\phi}_{s}$, making the product deployment of avatar stylization prohibitively computationally expensive.

With \name, we want to learn a model that can quickly adapt to arbitrary novel styles. 
Specifically, our goal is to find a set of suitable weight initialization $\phi_{meta}$, instead of initializing with $\phi$, that can approach $\hat{\phi}_{s}$ for unseen novel styles with a small number of $K$ steps (Figure \ref{fig:teaser}).

We leverage the idea of Reptile \cite{reptile}, a model-agnostic meta-learning method, to learn the initialization.
Reptile uses a bi-level training approach that trains on a distribution of tasks (in our case, a dataset of styles $\text{D}_\text{style}$),
and produces a fast learner that can adapt to a new style within a small number of update steps.

\begin{algorithm}
\caption{$\name$ for fast style adaptation}
\label{algo:Training}
%\SetAlgoNoLine
\KwIn{A pre-trained photorealistic Instant Avatar with parameters $\phi$, style dataset $\text{D}_\text{style}$, inner loop learning rate $\alpha$, outer loop learning rate $\beta$, meta batch size $M$, number of training iterations $T$, number of inner loop iterations $K$.}
\KwOut{Trained parameters $\phi_\text{meta}$ for fast style adaptation}
\begin{algorithmic}[1]
\STATE{Initialize $\phi_{\text{meta}_{0}}$ with $\phi$.}
\FOR{each iteration t = 1,..., T}
\FOR{each style index i = 1,..., M}
\STATE{$\phi_{s_i}^{0}$ = \textbf{Clone}($\phi_{\text{meta}_{t-1}}$).}
\STATE{Sample random style $s_i$ from $\text{D}_\text{style}$.}
\FOR{each iteration k = 1,...,K}
\STATE{$\hat{\phi}_{s_i}^{k} = \hat{\phi}_{s_i}^{k-1} - \alpha \nabla_{\phi}\mathcal{L}_{style}(\mathcal{M}_{\phi_{s_i}}, s_i)$}
\ENDFOR
\ENDFOR
\STATE{$\phi_{\text{meta}_{t}} = \phi_{\text{meta}_{t - 1}} + {\beta}\frac{1}{M}\sum^M_{i=1}(\phi^{K}_{s_i} - \phi_{\text{meta}_{t - 1}})$}
\ENDFOR
\end{algorithmic}
\end{algorithm}

Algorithm \ref{algo:Training} describes the outline of the meta-learning stage of $\name$.
In particular, in the inner loop, we perform stylization with SGD for $K$ steps to find a set of weights $\phi_{s_i}$ that minimize $\mathcal{L}_\text{style}$ for a particular style $s_i$, similar to Equation \ref{eq:SGD}.
In the outer loop, we aggregate the weights $\phi_{s_i}^K$ from style-specific models across many styles to update the weights of the meta model $\phi_{\text{meta}}$.
In practice, during training, to update $\phi_{\text{meta}}$, we use a meta-batch of styles with size $\text{M}$ instead of a single style, which has been shown to lead to faster learning due to variance reduction \cite{reptile}.
After the meta training stage, the meta-learnt model eventually serves as the initialization for fast adaptation to novel styles. 

It is worth highlighting the difference between using Reptile as described in Algorithm \ref{algo:Training} compared to regular training using SGD to minimize the expected loss across all styles in $\text{D}_\text{style}$.  
When $K=1$, Algorithm \ref{algo:Training} is indeed the same with joint-training on a mixture of all tasks.
\begin{equation}
\label{eq:grad_expectation}
\begin{aligned}
    \mathbb{E}_{s_i \in \text{D}_\text{style}}[\nabla_{\phi_{s_i}}\mathcal{L}_\text{style}(\mathcal{M}_{\phi_{s_i}}, s_i)] = \nabla_{\phi_{s_i}}\mathbb{E}_{s_i \in \text{D}_\text{style}}[\mathcal{L}_\text{style}(\mathcal{M}_{\phi_{s_i}}, s_i)]
\end{aligned}
\end{equation}
However, when we perform multiple gradient updates ($K>1$), the expected update is no longer equal to update on the average function, since the expected update depends on the higher order derivative of $\mathcal{L}_\text{style}(\mathcal{M}_{\phi_s}, s_i)$. 
For each individual's avatar network, we train a meta-learnt model to find a good initialization for each specific avatar.
As shown in Figure \ref{fig:network}, we only update the bias mappers, and keep the rest of the Instant Avatar network frozen.

Notably, after the meta training stage, our model does not produce the final set of weights that is "optimal" for any specific style.
Instead, our model produces a suitable initialization that can quickly adapt to new styles.
This is different from training hypernetworks \cite{ha2017hypernetworks} to directly predict a new set of weights corresponding to every new style in a single, feed-forward pass. 
These models can add a significant number of parameters to the stylization model, making it non-trivial to train, and sometimes leads to less competitive results \cite{chiang2021stylizing, hyperdomainnet,8578939}.

\section{Results}
\begin{figure*}
\centering
	\includegraphics[width=0.95\linewidth]{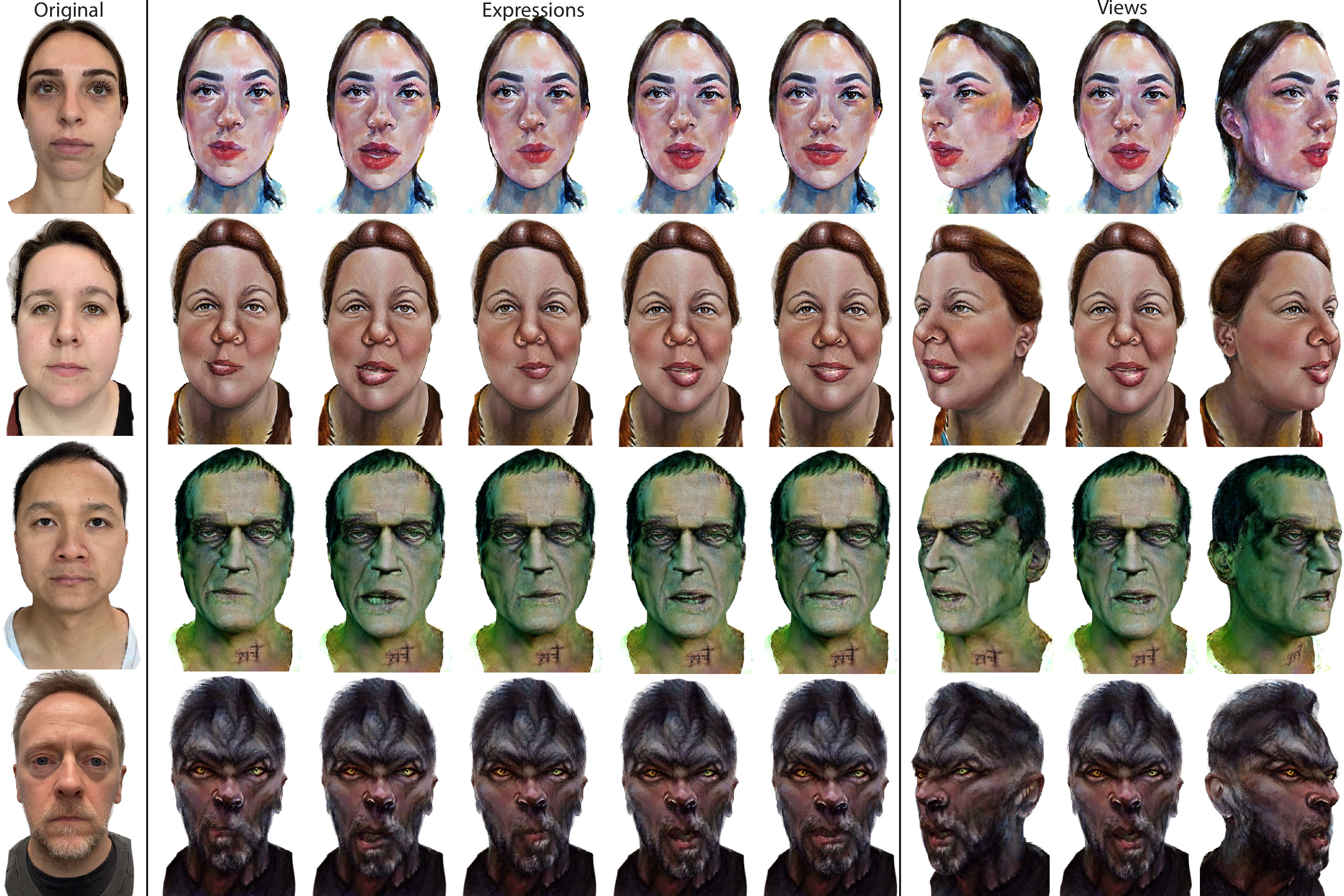}
	\caption{\label{fig:expressions_views}
		\textbf{Qualitative results by $\name$}. Here we show the stylized results by $\name$ in a variety of identities, styles, expressions and views. Using only text descriptions (from top to bottom: \textit{"Gouache painting"}, \textit{"Caricature"},  \textit{"Frankenstein"},\textit{"Werewolf"}) as our style guidance, we show that $\name$ can generate consistent stylized results across different expressions and views.}
\end{figure*}

\subsection{Implementation Details}
\subsubsection{Training with CLIP}
We use pre-trained CLIP models \cite{CLIP}, in particular, both CLIP-ViT/B-32 and CLIP-ViT/B-16 to extract the image statistics.
We find that image and text augmentation play an important role in the amount of stylization effect when using CLIP features, as also observed in VQGAN-CLIP \cite{VQGAN-CLIP} and PureCLIPNeRF. \cite{pureclipnerf}.
We use the same augmentation strategies from Tex2Live \cite{bar2022text2live} for images and from StyleGAN-NADA \cite{stylegannada} for texts.
\subsubsection{Training details}
We choose to stylize a diverse set of avatars that spans a wide range of visual attributes, such as skin color or hairstyle, and includes challenging features such as facial hair or a beard.
We render images at size 1920 $\times$ 1080 of the avatar with random views and expressions.
In particular, we render 2 side views and 2 frontal views, with small random perturbations to each view along the azimuth.
We set $\lambda_{\text{CLIP}}=1.0$, $\lambda_{\text{ID}}=1.0$.
For the style dataset $\text{D}_\text{style}$ for the meta-learning training phase, we generate 500 texts describing the appearance of either real humans (\textit{"A person with dark hair and a faint scar above their left eyebrow."}) or fictional identities (\textit{"Woodland elf with heart-shaped face, soft features, and a button nose and warm brown eyes"}), as well as makeup and dress-up descriptions (\textit{"Dark angel with black wings eyeliner and smoky makeup"}).
For the test style dataset in section \ref{sec:meta_compare}, we only use art styles that have not been used during the meta-training stage, such as \textit{"Watercolor painting"} or \textit{"bronze sculpture"}. We include a more extensive list of examples from the dataset in the supplementary document.
We use a meta learning rate of 6e-3 and the inner learning rate of 4e-3.
For each style, for the inner loop, we perform stylization by minimizing $\mathcal{L}_\text{style}$ for 200 steps.
For the fast adaptation stage, starting from the learnt initialization, we run stylization for 200 steps with the learning rate of 8e-3.
We train and evaluate our models on an NVIDIA V100 GPU with 32 GB of memory.
The meta-training stage takes 16 hours on average for each avatar and once trained, each avatar can adapt to a new style in as quickly as less than two minutes.
\subsection{Comparisons}
In Section \ref{sec:qualitative} and \ref{sec:quantiative}, we compare to a range of text-guided portrait stylization methods, including 2D (StyleGAN-NADA \cite{stylegannada}, HyperDomainNet \cite{hyperdomainnet}) and 3D approaches (DATID-3D \cite{datid3d}).
We could not compare with RODIN \cite{rodin} and recently published CLIPFace \cite{clipface} due the the lack of publicly available datasets, pre-trained models and code.

Note that StyleGAN-NADA, HyperDomainNet and DATID-3D rely on pre-trained image generative models for faces \cite{Karras_2019_CVPR}.
For StyleGAN-NADA, we train the model to adapt to the new style domains using the author-provided code and settings.
For HyperDomainNet and DATID-3D, we use the author-provided code and pre-trained models.
In the supplementary material, we also include a comparison with VToonify \cite{yang2022Vtoonify}, a video portrait stylization approach.

%%%%%%%%%%%%%%%%%%%%%%%%%%%%%%%%%%%%%%%%%%%%%%%%%%%%%%%%%
%%%%%%%%%%%%%%%%%%%%%%%%%%%%%%%%%%%%%%%%%%%%%%%%%%%%%%%%%
%%%%%%%%%%%%%%%%%%%%%%%%%%%%%%%%%%%%%%%%%%%%%%%%%%%%%%%%%
\begin{figure}
\includegraphics[width=\linewidth]{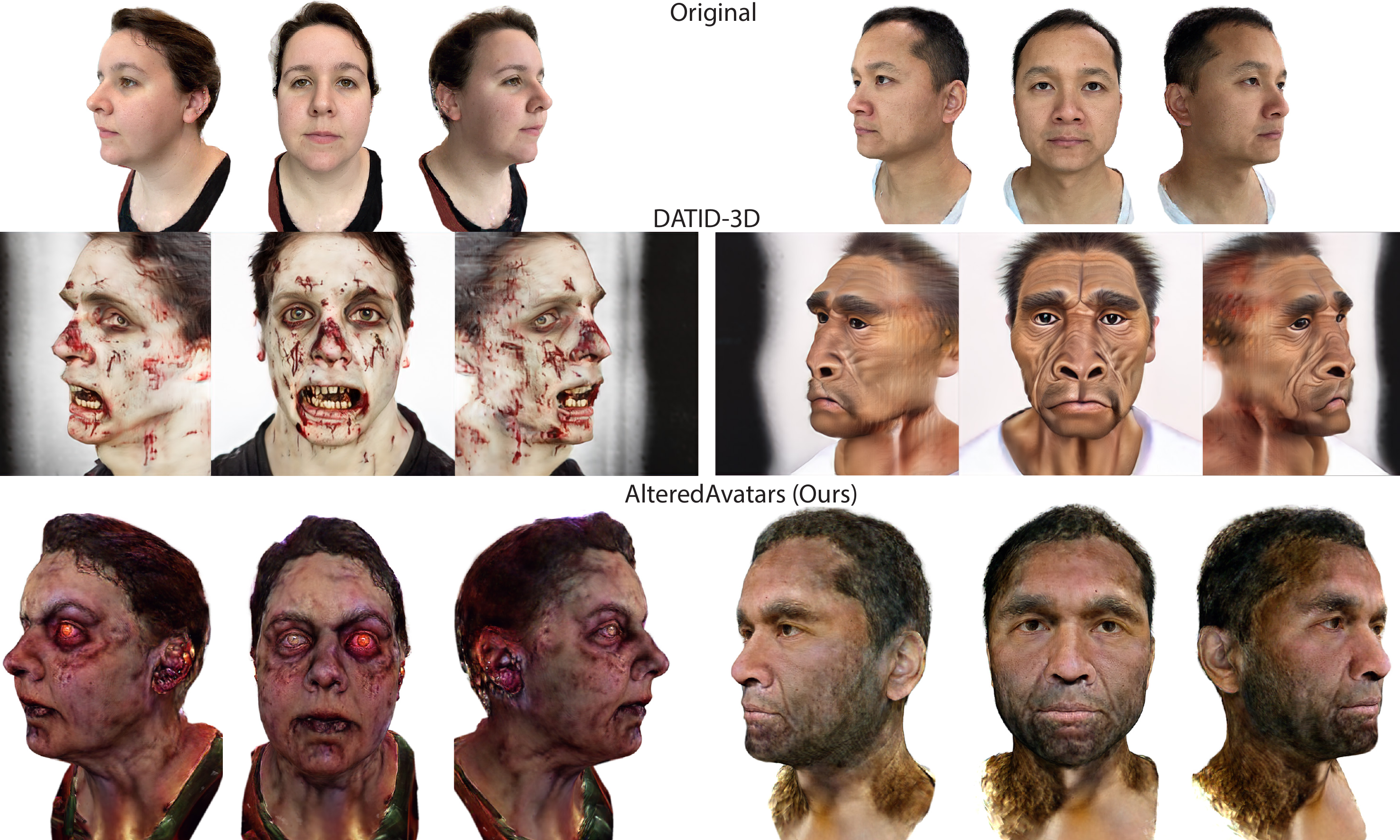}
\caption{\label{fig:qualititve_comparison_3d}
\textbf{Qualitative comparison for text-guided 3D avatar stylization}. We show stylized results for two styles \textit{"Zombie"} (left) and \textit{"Neanderthal"} (bottom). Only $\name$ can maintain the avatars' original identity, expressions, pose and eye gaze well.}
\end{figure}

\subsection {Qualitative comparison}
\label{sec:qualitative}
\begin{figure}
\centering
\includegraphics[width=\linewidth]{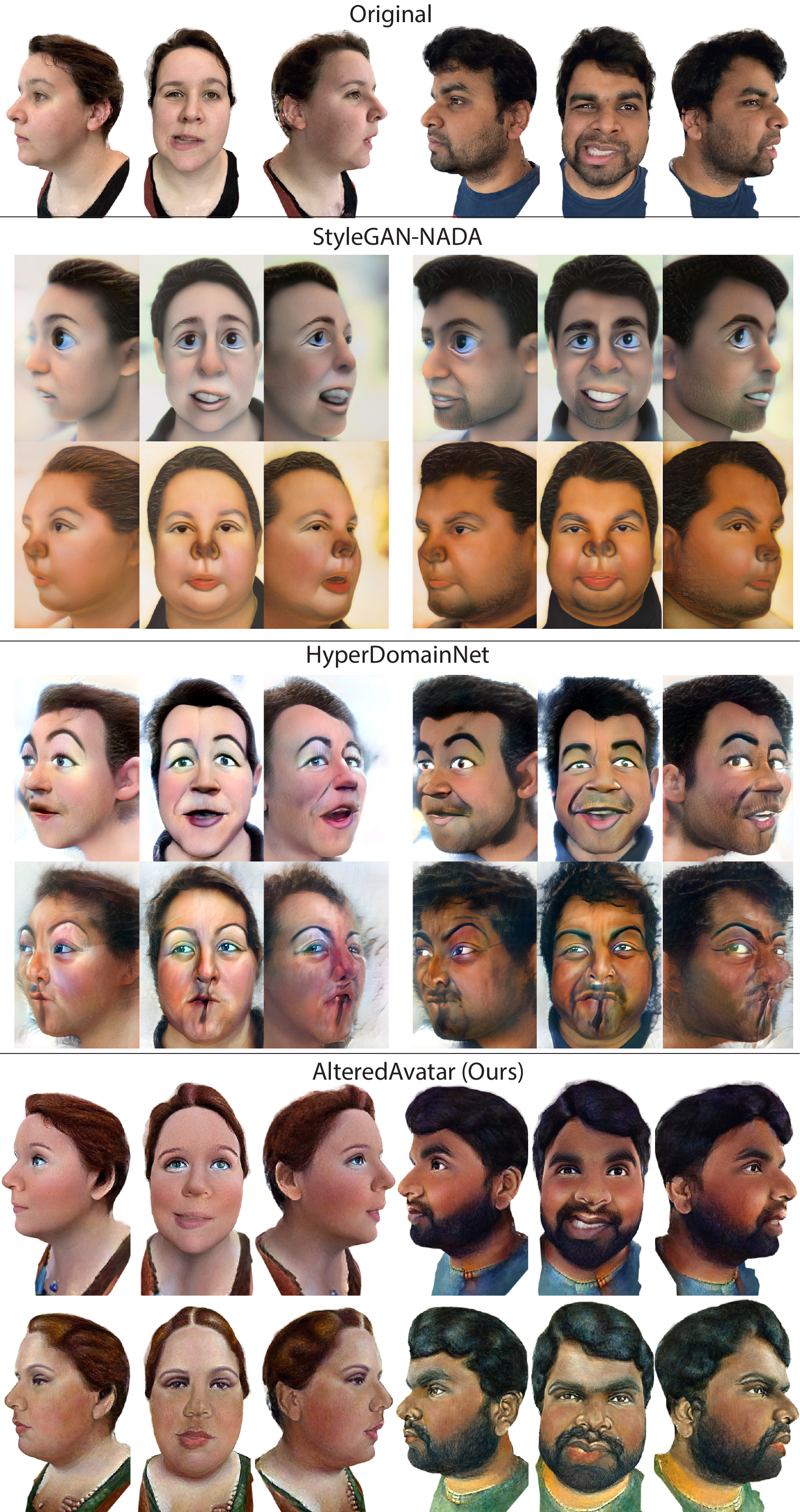}
\caption{\label{fig:qualititve_comparison_image}
\textbf{Qualitative comparison for image stylization}. We show stylized results for two styles \textit{"3D render"} (top) and \textit{"Fernado Botero painting"} (bottom). Only $\name$ can maintain the avatars' original expressions, pose and eye gaze well.}
\end{figure}
\begin{figure*}
\includegraphics[width=\linewidth]{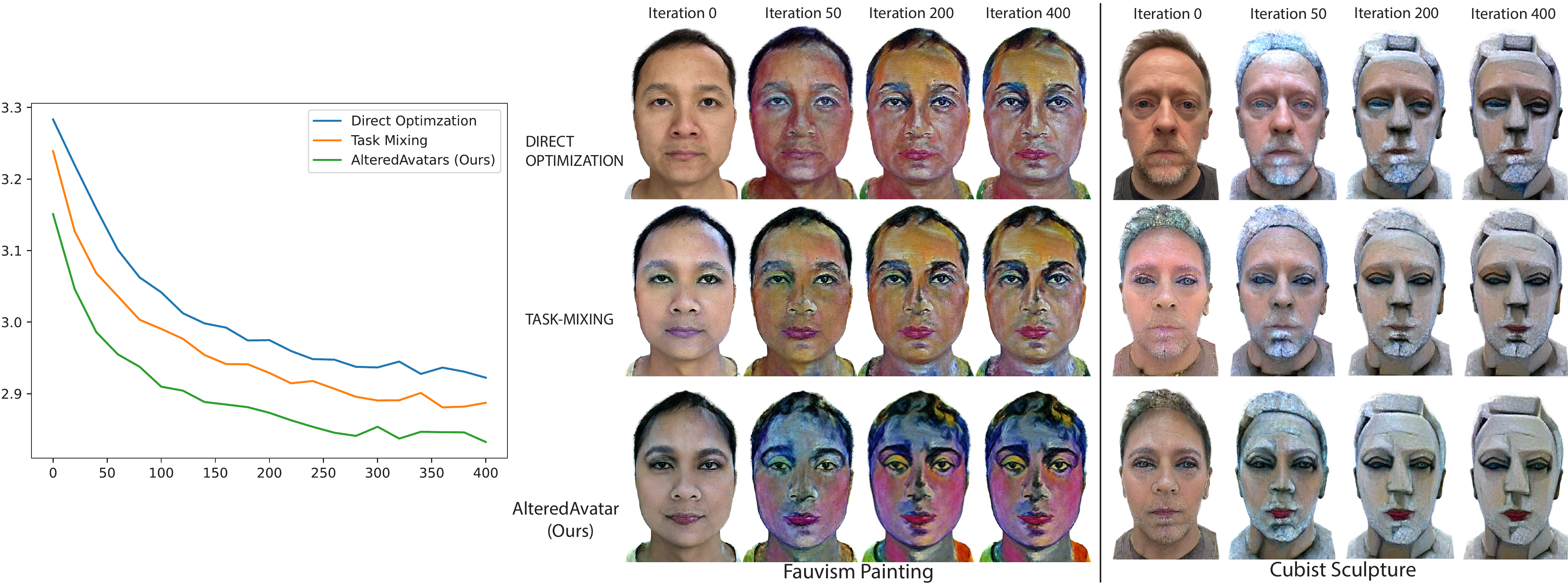}
\caption{\label{fig:qualititve_comparison_meta}
\textbf{Effects of using different initializations.}Comparison between different initializations: the original Instant Avatar weights (\textsc{Direct optimization} - green line), a learnt initialization from a task-mixing model (\textsc{Task-mixing} - blue line), and meta-learnt initialization ($\name$- orange line). After only 50 iterations, $\name$ already produces results that capture the target style well, while \textsc{Direct Optimization} and \textsc{Task-mixing} only start to add some style effects to the avatars. After convergence, $\name$ also achieves the lowest $\mathcal{L}_\textit{style}$, which is computed across 25 novel art styles and 5 identities.}
\end{figure*}

Figures \ref{fig:teaser} and \ref{fig:expressions_views} show qualitative results by $\name$ on different avatars, views and expressions. $\name$ persistently stylizes and preserves a suitable amount of the avatars' original identities, despite the large variety in their visual attributes. More importantly, although $\name$ only uses the style guidance from a single text description, our stylized avatars are consistent across different views and expressions.

It is worth highlighting each StyleGAN-NADA model is trained to work with a single style, whereas HyperDomainNet is a hypernetwork trained with 1020 texts created by sampling CLIP embeddings from a convex hull of the initial 52 text embeddings.
In contrast, $\name$ only needs a single text or single image for the novel style adaptation stage, after the meta-training stage with 500 styles.

As shown in Figure \ref{fig:qualititve_comparison_image}, image-based methods (StyleGAN-NADA and HyperDomainNet) cannot handle more extreme views, such as left and right side view, due to their reliance on StyleGAN \cite{Karras_2019_CVPR}, which was trained mostly on frontal or near-frontal views. %
They also tend to modify the avatars' eye gaze and expressions.
Since these two methods fail to generate stylized results from side views, we only show the closest side views with which we can run the models.
Additionally, HyperDomainNet seems to have style mode collapse, and thus generates similar-looking stylized results even with different style inputs.
$\name$ on the other hand can handle a wide range of views including very left and right sides, and maintains the avatars' original expressions and eye gaze.

Figure~\ref{fig:qualititve_comparison_3d} shows a comparison between $\name$ and DATID-3D \cite{datid3d}, which uses a learnt tri-plane representation. Note that DATID-3D can only generate static avatars. However, even with the learnt 3D representation, DATID-3D still struggles with more side views beyond the frontal view, similar to StyleGAN-NADA and HyperDomainNet, and generates distortion artifacts. DATID-3D also significantly alters the original avatars' identities and expressions, which is undesirable for MR/VR applications. 
\subsection {Quantitative comparison}
\label{sec:quantiative}
\subsubsection{Semantic score}
\label{sec:blip}
% Table
\begin{table}
\caption{\textbf{Qualitative comparisons for semantic score}. We compare the BLIP score (the higher the better) that measures the semantic similarity scores between rendered images of the stylized avatars and the target style. The best result is in bold.}
\label{tab:blip}
\begin{minipage}{\columnwidth}
\begin{center}
\begin{tabular}{*{2}{c}}
  \toprule
  Methods & BLIP\\ \midrule
  StyleGAN-NADA \cite{stylegannada} & 0.36 $\pm$ 0.03\\
  HyperDomainNet \cite{hyperdomainnet} & 0.33 $\pm$ 0.04\\
%   \midrule
%   VToonify \cite{yang2022Vtoonify} & 0.34 $\pm$ 0.04\\
  \midrule
  DATID-3D \cite{datid3d} & 0.33 $\pm$ 0.04\\
  \midrule
  $\name$  (Ours) & \textbf{0.41 $\pm$ 0.03}\\
  \bottomrule
\end{tabular}
\end{center}
\bigskip\centering
\end{minipage}
\end{table}%

Here we estimate the semantic similarity between the stylized avatars and the target styles.
For example, we want to measure how closely an avatar stylized into a zombie style matches the text description \textit{"a zombie"}.
Essentially, this measures how well results from each method can match the target styles.

All methods, including ours, use the style guidance from CLIP during training.
Therefore, we use the embeddings of BLIP \cite{blip}, instead of CLIP, to estimate the semantic score as follows:
\begin{equation}
 \mathcal{S}_\text{BLIP}=\langle\, \mathcal{E}_{i}({\mathbf{I}_{tgt}}), \mathcal{E}_{t}(\mathbf{t}_{tgt})\rangle.
\end{equation}
where $\mathcal{E}_{i}$ and $\mathcal{E}_{t}$ denote BLIP pre-trained image and text encoder, respectively.
Similar to CLIP, BLIP is a pre-trained multi-modal model for vision and language, and thus also learns an image-text common embedding space. 
However, BLIP is trained on image and caption pairs, and on additional tasks to image-text contrastive learning.
We compute the BLIP score using five styles with five different avatars rendered from random views and report the mean and standard deviation of the BLIP score in Table \ref{tab:blip}. $\name$ achieves the highest semantic score, showing that its stylized results match the target style the closest.
\subsubsection{User study}
\label{Sec:user_study}
\begin{figure}
\includegraphics[width=\linewidth]{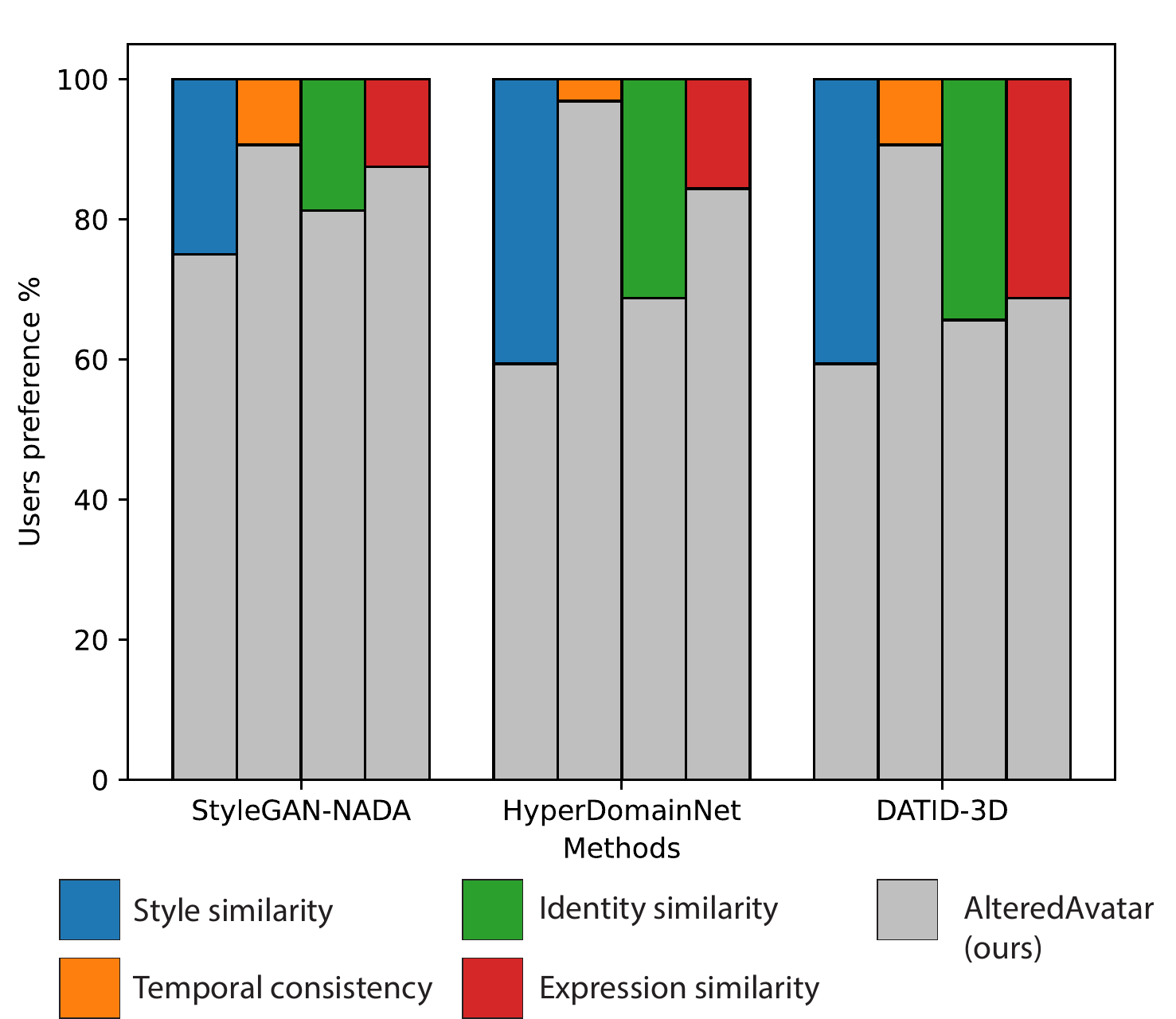}
\caption{\label{fig:User_study}
\textbf{User study}. We present two videos of novel view synthesis results, one generated by our method (grey) and one by another approach. We ask the participant to select the one that (1) Matches the target style better (blue) (2) Has less temporal inconsistencies (orange) (3) Preserves more identity of the original avatar and (4) Preserves more expressions of the original avatar.}
\end{figure}
In addition to measuring the semantic score between the stylized results and the style descriptions in Section \ref{sec:blip}, we conduct a user study to measure users' perception of the results by $\name$ and three alternative approaches (StyleGAN-NADA \cite{stylegannada}, HyperDomainNet \cite{hyperdomainnet}, and DATID-3D \cite{datid3d}).
In particular, we want to measure users' judgement on 4 separate aspects: 1) Which method matches the input style better; 2) Which method generates smoother results without temporal inconsistencies; 3) Which method generates results that match the original avatars' expression; and 4) Which method generates results that preserves the original avatars' identities better. 
For each question, we ask the participants to compare two videos of two stylized avatars with the same identity and style, one generated by $\name$ and the other by one alternative method.
To generate results for the study, we pick five different avatars with three different styles, and render them with different expressions while rotating the views along the azimuth.
We collect answers from 16 participants, and present the results in Figure \ref{fig:User_study}.
For every comparison aspect, we plot the percentage of users choosing our method (colored in grey in Figure \ref{fig:User_study}) over the method against which we are comparing.
Our method achieves much better results in all 4 categories, even on temporal consistency when compared with a 3D approach like DATID-3D.
\begin{figure}
\includegraphics[width=\linewidth]{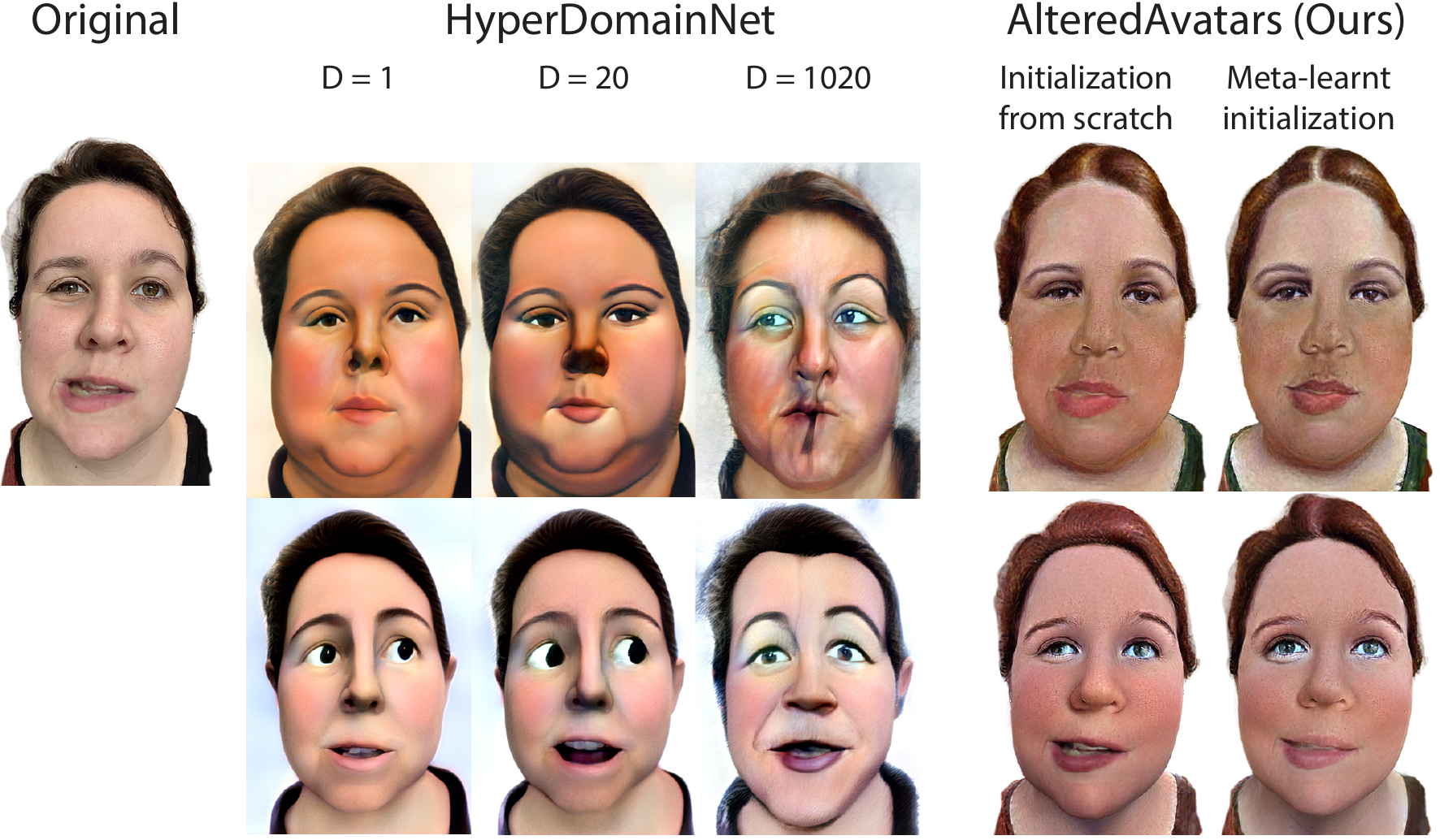}
\caption{\label{fig:discussion_hypernet}\textbf{Hypernetworks vs. Fast Adaptation with Meta-learnt Initialization}. HyperDomainNet trains a feed-forward stylization network for arbitrary stylization. As the size of the data set increases ($D=1$ (overfit to one style) to $D=1040$), the quality of the stylized results drops significantly. Meanwhile, $\name$ learns an initialization that can adapt using a small number of update steps. Therefore, the quality between the direct optimization to approach and $\name$ are comparable, with $\name$ using less time for stylization. We show stylized results for two styles \textit{"Fernado Botero painting"} (top) and \textit{"3D cartoon render"} (bottom).}
\end{figure}
\subsection {Fast style adaptation}
\label{sec:meta_compare}
We compare our approach for fast style adaptation with two alternatives: \textsc{direct optimization} and \textsc{task-mixing}.  
For $\name$, we perform stylization to a novel style using our meta-learnt initialization.
For \textsc{direct optimization}, we perform stylization starting from the weights of the original photorealistic avatars.
For \textsc{task-mixing}, we perform stylization using the initialization from a joint-training model that is trained to minimize $\mathcal{L}_{style}$ over all $\mathcal{D}_{style}$.
This model was trained in the same manner with $\name$ (Algorithm \ref{algo:Training}) with the same style dataset $\mathcal{D}_{style}$, but using inner step $K=1$, meta batch size $M=1$, and training for 100000 iterations.
For all three approaches, we compute the average loss across 25 novel art styles that were not included in the meta-training dataset $\mathcal{D}_{style}$.
We run stylization for 400 steps for all models.

Figure \ref{fig:qualititve_comparison_meta} shows stylizing using our meta-learnt initialization leads to smaller $\mathcal{L}_\text{style}$ in a shorter amount of time.
After 50 steps, \textsc{Direct Optimization} only starts to apply some style appearance to the avatar, whereas $\name$ can already produce well-stylized results.
\textsc{Task-mixing} applies more style effects to the avatars compared to \textsc{Direct Optimization}, but still lags behind $\name$.
Compared to the baseline \textsc{Task-mixing}, $\name$ has learnt an effective initialization that enables fast adaptation to novel styles.

When it comes to handling novel styles for stylization, we want to highlight the advantage of learning a good initialization for fast adaptation, over using a universal one-shot style-transfer network. 
This is the approach taken by HyperDomainNet, which is trained on a large dataset of styles and does one-shot generalization to a new style without any fine tuning. 
As shown in Figure \ref{fig:discussion_hypernet}, HyperDomainNet produces very convincing results when trained to overfit to a single domain, or even to 20 domains.
However, it struggles to generalize to a potentially arbitrary unlimited domain.
Note that both styles \textit{"Fernado Botero painting"} and \textit{"3D render style"} shown here are in the training dataset for the multi-domain version of HyperDomainNet, yet the results are less competitive than its single-domain version and $\name$.
In other words, there is a considerable gap between the version overfit to a single-domain and the multi-domain version of HyperDomainNet.
We argue that it is extremely challenging to train a feed-forward model that can both generalize to an arbitrary style and maintain the high fidelity requirement of avatars.
Instead, we opt to learn a good initialization that allows for fast stylization within a few update steps, achieving comparable results to direct optimization approaches as shown in Figure \ref{fig:qualititve_comparison_meta} and \ref{fig:discussion_hypernet}.
We believe that with $\name$, we can strike a good balance between the speed, flexibility to adapt to new styles, and quality.
%%%%%%%%%%%%%%%%%%%%%%%%%%%%%%%%%%%%%%%%%%%%%%%%%%%%%%%%%
%%%%%%%%%%%%%%%%%%%%%%%%%%%%%%%%%%%%%%%%%%%%%%%%%%%%%%%%%
%%%%%%%%%%%%%%%%%%%%%%%%%%%%%%%%%%%%%%%%%%%%%%%%%%%%%%%%%
%\end{figure}
\subsection {Additional experiments}
\label{Sec:Ablation}
\begin{figure}
\includegraphics[width=0.96\linewidth]{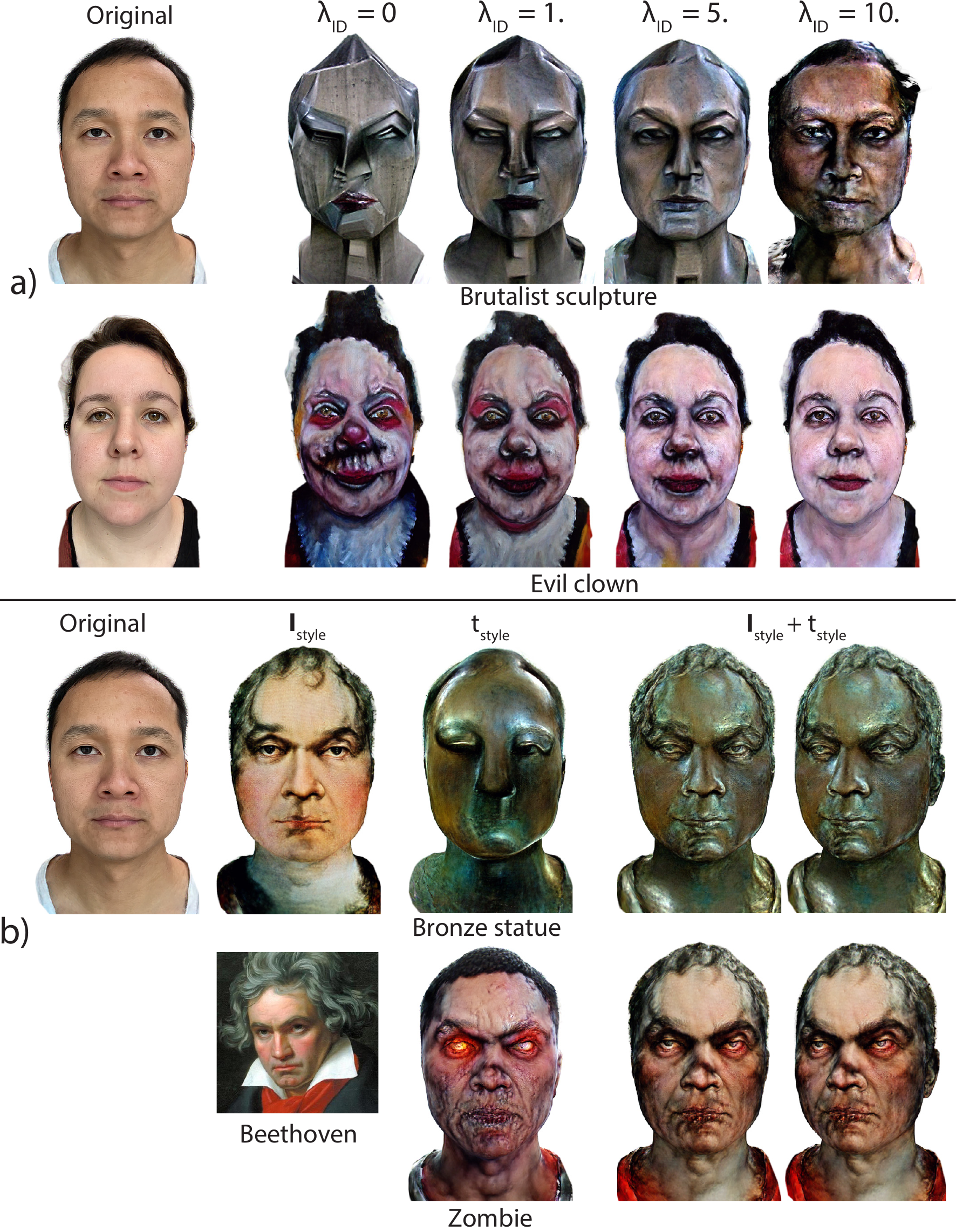}
\caption{\label{fig:style_mixing}\textbf{Top-a}.The strength of stylization effects progressively reduce as we increase $\lambda_\text{ID}$.
\textbf{Bottom-b}. Stylization results using only a style image (second column), style text description (third column), and a combination of both (fourth column).}
\end{figure}

%\paragraph{Ablation} \textbf{Identity preservation.}
\paragraph{Identity preservation}
$\name$ offers control over how much stylization is applied to the avatar with the identity loss $\mathcal{L}_\text{ID}$.
As shown in Figure \ref{fig:style_mixing}-a, small values of $\lambda_\text{ID}$ result in significant changes in the geometry and appearance of the avatar. 
Meanwhile, larger values of $\lambda_\text{ID}$ preserve more of the original facial structure and appearance of the avatars.
The amount of desirable stylization greatly depends on personal preference and the chosen styles, hence we use $\lambda_\text{ID}$ as a tuning knob to control the amount of preservation of the original identity.
In this paper, we use $\lambda_\text{ID}=1.0$.
\paragraph{Style mixing}
By using CLIP features for stylization, $\name$ enables users to change the style of avatars using an image, a text description, or both, opening up potentials for creating new unique avatars.
Figure \ref{fig:style_mixing}-b shows novel styles can be created with $\name$ using both target style images and texts. 

In the supplementary document, we include an ablation study for the Instant Avatar architecture for stylization, and visualizations of the geometry of the stylized avatars.

\section{Conclusion}
We introduce $\name$, a meta-learnt avatar representation that can quickly adapt to novel styles using a smaller number of update steps, offering a balance between speed, flexibility, and quality.
The stylized avatars can be driven to different expressions, and maintain consistent appearance across views and expressions.

Our method presents a few limitations. For example, the method does not model extra geometry for accessories that could be a strong indicator of styles, such as a witch wearing a hat or a devil with two horns. As our model uses global stylization, it does not support styles such as "add a mustache" or "afro hair style" that only change parts of the avatar. Additionally, as currently there is no disentanglement between the shading and albedo in the original avatar, shading effects might be exaggerated in the stylized results, leading to unnatural darkening face regions. We also notice some expression dampening effects in the stylized avatars compared to the original ones, which is partly due to the entanglement of expression and identity in the Instant Avatar representation, as also noticed by Cao {\emph{et al.}} \cite{ICA}.
Finally, the quality of our stylized avatars depends on the quality of the photorealistic Instant Avatar. 
We include some examples of the failure cases in the supplementary material.

% Bibliography
{\small
\bibliographystyle{ieee_fullname}
\bibliography{ref}
}
\clearpage
\appendix
\newpage
\section{Where to fine-tune}
\begin{figure}
	\includegraphics[width=\linewidth]{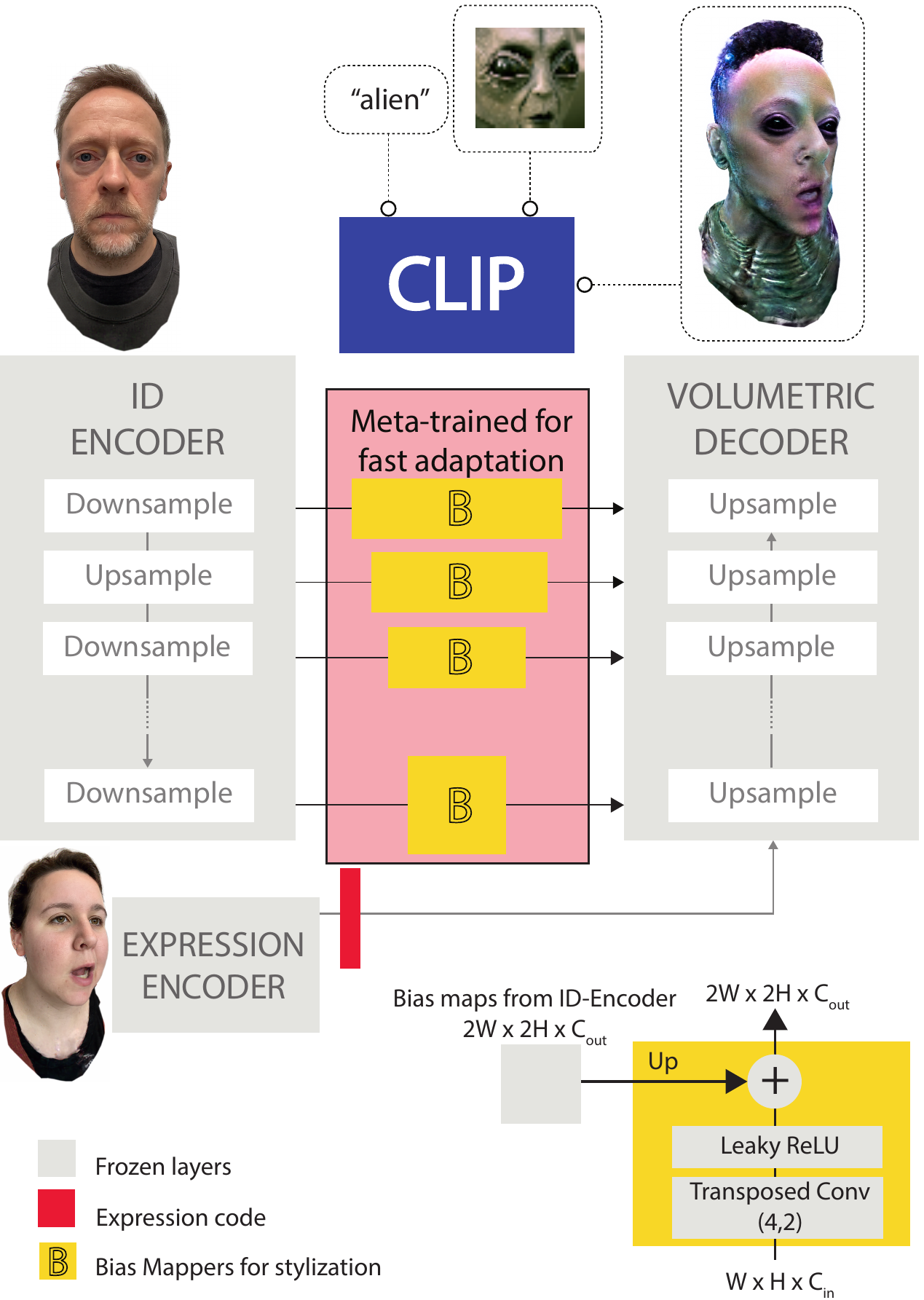}
	\caption{\label{fig:network2}
		\textbf{Overall pipeline for $\name$}. Taking the Instant Avatar~\cite{ICA} as the avatar representation, we only use the weights of its bias mappers (yellow blocks) during the stylization process, and keep the rest of the network (the remains of the Identity Encoder, Expression Encoder, and Volumetric Decoder) frozen. Rendered images of the stylized avatar (from different views and expressions) are given to a pre-trained CLIP model to compute the stylization loss $\mathcal{L}_\text{style}$ against an input style text description or a target style image. The stylized avatar can be driven using expressions from the same or from another avatar.}
\end{figure}
\begin{figure}
	\includegraphics[width=\linewidth]{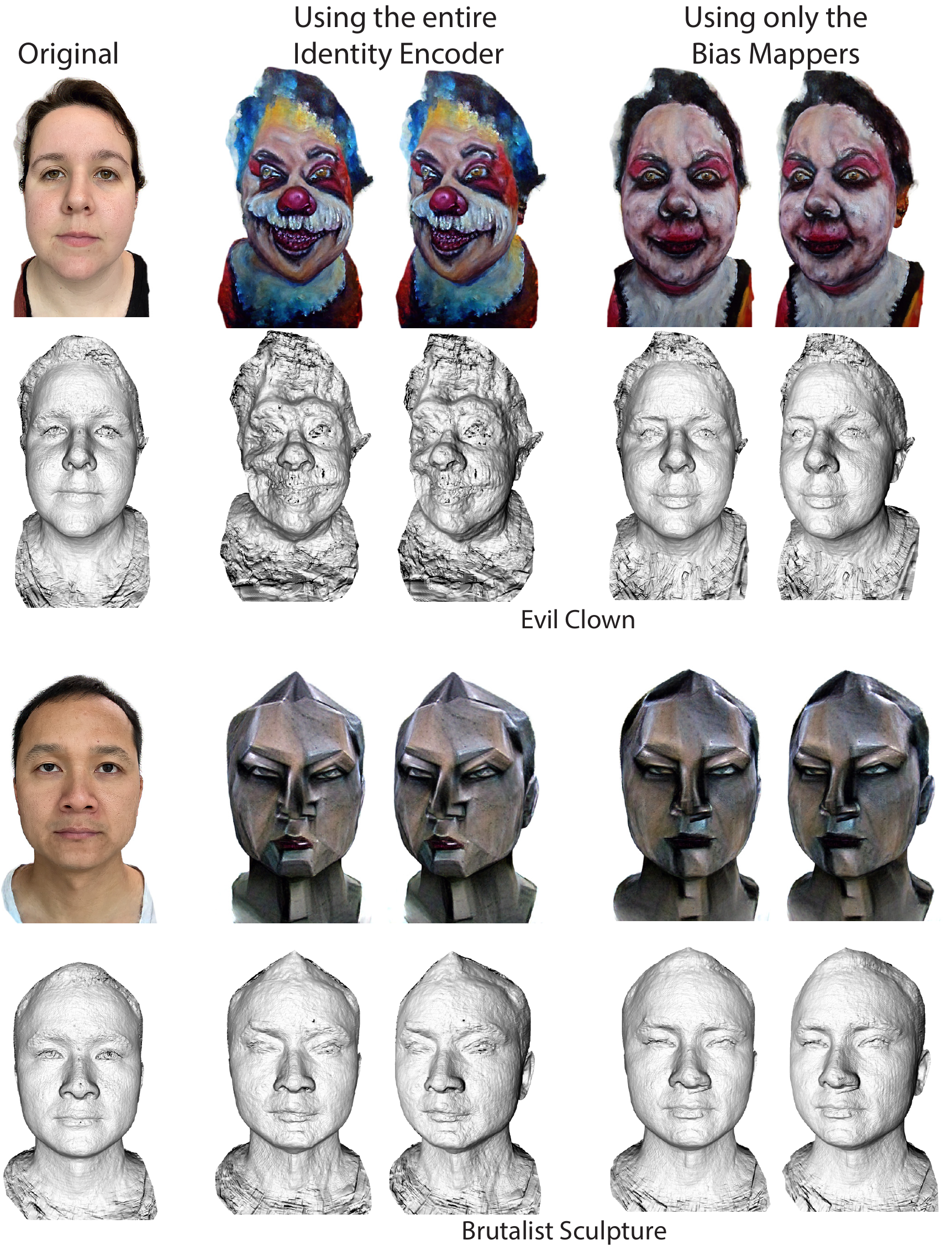}
	\caption{\label{fig:where_to_stylize}
		\textbf{Network architecture and stylization effects}. Using only the bias mappers in the Identity Encoder achieves the same level of stylization effect (top) and avoids excessive geometry distortion during stylization (bottom).}
\end{figure}

One key question when we adopt a learnt avatar representation like Instant Avatar \cite{ICA} is which part of the network is suitable for high-quality stylization that is consistent across views and expressions.
While a straightforward choice is to use the weights of the Identity Encoder, we find using all the layers in this encoder redundant.
Particularly, we find that using only the blocks of bias mappers of the Identity Encoder (see Figure \ref{fig:network2}) is enough to achieve a good level of stylization effects.
Changing fewer parameters also implicitly acts as a regularizer to the avatar network, which helps maintain the overall facial structure and avoid overly-distorted geometry.
Notably, with more parameters to update when using the entire Identity Encoder, we have to reduce the learning rate significantly from 8e-3 to 1e-3 to avoid training collapse, lengthening the stylization process.
As shown in Figure \ref{fig:where_to_stylize}, changing only the bias mappers, we can achieve comparable stylization results with using all the layers in the Identity Encoder and avoid excessive geometry changes.

\section{Geometry stylization}
\begin{figure}
	\includegraphics[width=\linewidth]{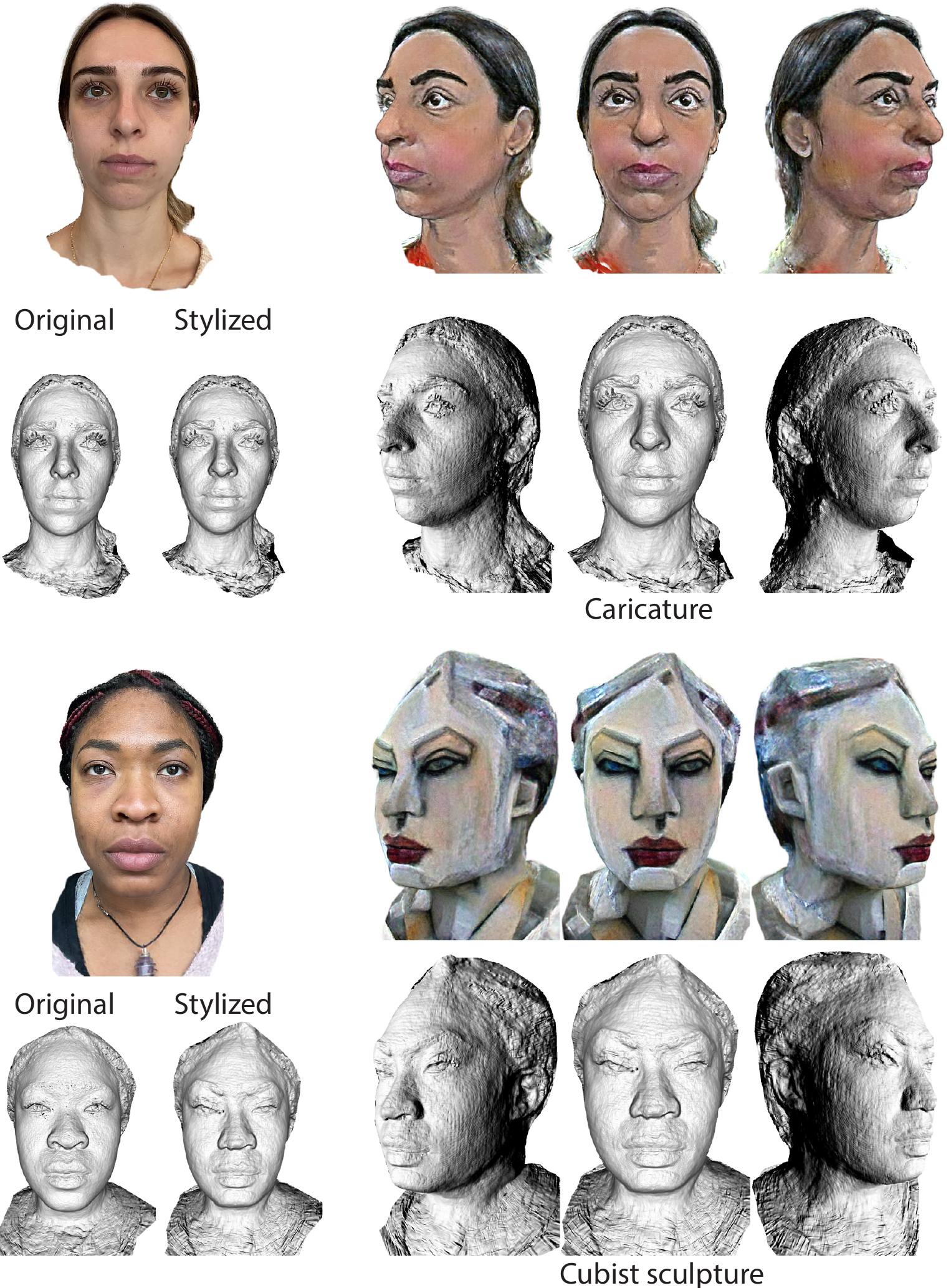}
	\caption{\label{fig:geometry_visual}
		\textbf{Geometry changes}. $\name$ stylizes both the appearance and geometry of the avatar to better capture the target style.}
\end{figure}
Instead of just changing the texture appearance to match a target style, $\name$ modifies both the avatar's appearance and 3D geometry, which has been shown to be an important factor of style \cite{Kim20DST, 9577906}.
Here, we visualize the geometry of the stylized avatars by ray-marching the avatar's underlying MVP representations.
As shown in Figure \ref{fig:geometry_visual}, the geometry of the stylized avatars changes to accommodate the target style compared to the original photorealistic avatars.
This ranges from small changes in the nose of the avatar for the style \textit{"Caricature"} (Figure \ref{fig:geometry_visual}-top), to larger changes in the overall geometry of the avatar for the style \textit{"Cubist sculpture"}(Figure \ref{fig:geometry_visual}-bottom).
\section{Failure case}
\begin{figure}
\centering
	\includegraphics[width=\linewidth]{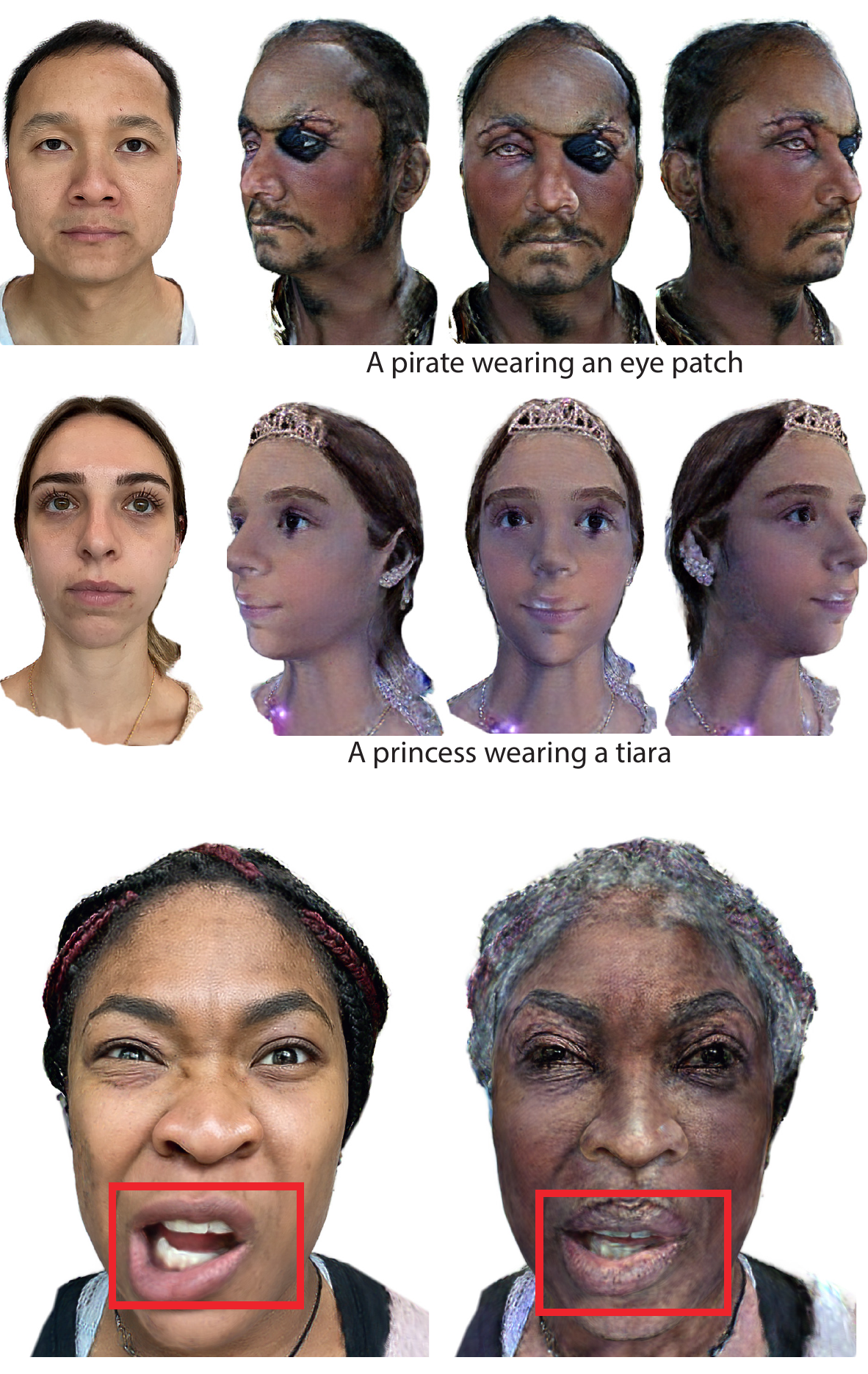}
	\caption{\label{fig:failure}
		\textbf{Failure cases}. \textbf{Top:} $\name$ currently cannot handle accessories such as an eye patch or a tiara, which can be a great indicator of style. \textbf{Bottom:} Blurriness in the teeth region (highlighted in the red box) is transferred from the original to the stylized avatar. The facial expression is also slightly dampened in the stylized avatar.}
\end{figure}
Figure \ref{fig:failure} shows examples for some current limitations of our method. Currently $\name$ struggles with modelling extra geometry accessories, which can be an important factor for styles. Additionally, the stylized results heavily depend on the quality of the original photorealistic avatar. As shown in Figure \ref{fig:failure}-bottom, a blurry artifact in the original avatar results in the blurry artifact in the stylized avatar. We also notice some expression dampening effects in the stylized avatar compared to the original ones (see Figure \ref{fig:failure}-bottom).
\section{Comparison with video stylization}
\begin{figure}[h]
	\includegraphics[width=\linewidth]{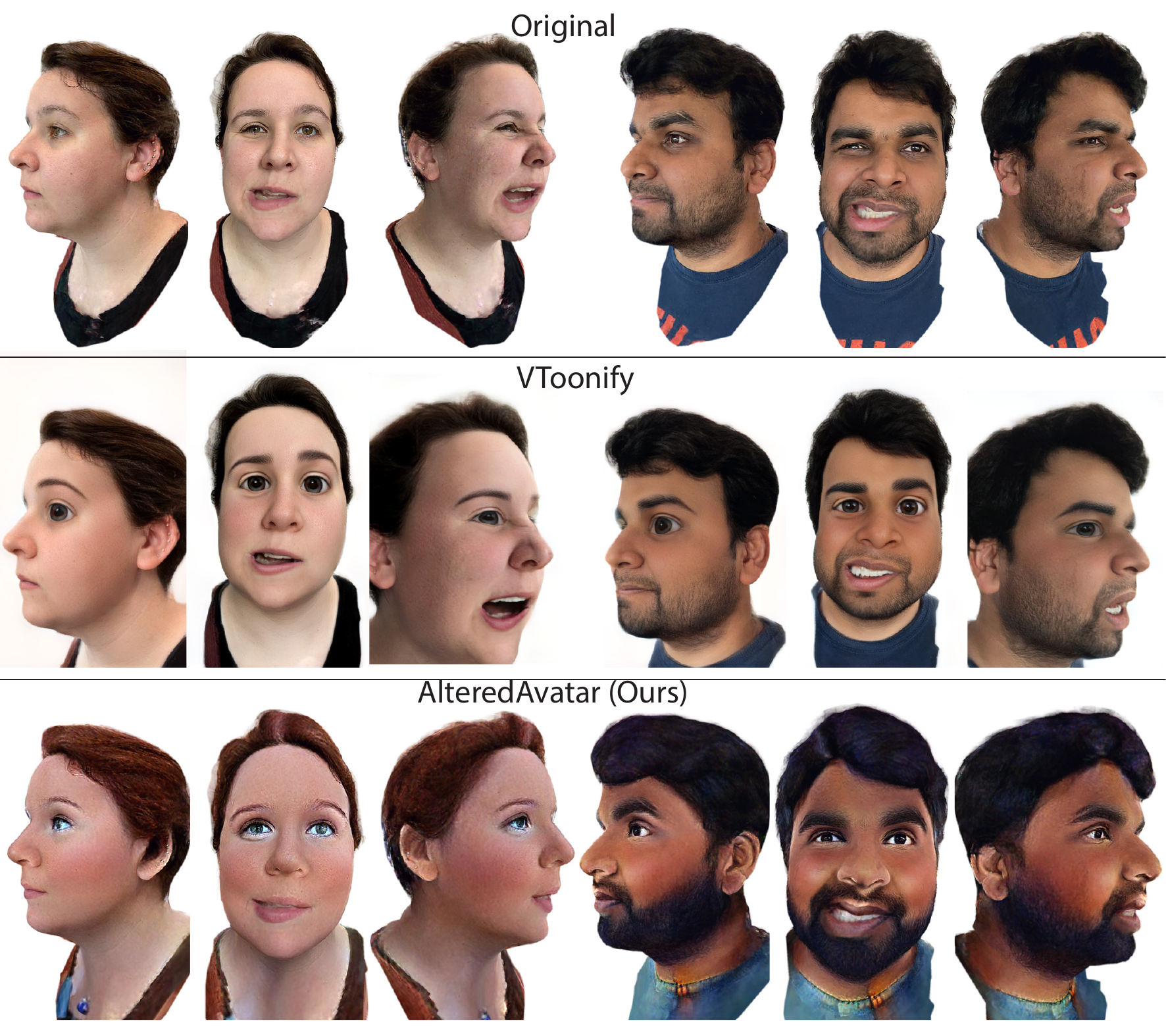}
	\caption{\label{fig:qualititve_comparison_video}
		\textbf{Qualitative comparison for video stylization}. We show stylized results for style \textit{"3D cartoon"}. Both methods produce comparable results, although VToonify is trained on a dataset of a few hundred images for every style, and \name only uses the style guidance from a single image or text description. Moreover, VToonify makes slight modifications to the avatars' eye gaze, and cannot handle more side views.}
\end{figure}

Figure \ref{fig:qualititve_comparison_video} shows results for video stylization, Vtoonify \cite{yang2022Vtoonify}.
VToonify shows very competitive results with $\name$ in terms of view, eye gaze and expression consistency.
However, it still cannot handle more extreme left or right views, and modifies the avatar's eye gaze in more side views.
Thus, here we only show the closest side views with which we can run VToonify.
More importantly, for every style, it requires a dataset of a few hundred reference images, and an additional process to create a paired training dataset by combining StyleGAN \cite{Karras_2019_CVPR} and DualStyleGAN \cite{dualstylegan}, making it non-trivial to adopt novel styles. 
These datasets are well-curated and even fine-tuned to each gender.
On the other hand, $\name$ only uses a text description as its style guidance.
The comparison therefore highlights the difference in the performance between one-shot avatar stylization using a single text description, and stylization with a well-curated dataset.
\pagebreak
\section{Dataset}
For the metra-training stage, we generate 500 text descriptions using ChatGPT \cite{chatgpt}. This includes descriptions for real humans across ages, gender and races. Additionally, we also use descriptions for fiction characters such as witches or fairies. Finally, we use descriptions for makeup and dress-up styles. A sample of the training dataset is listed below. Upon acceptance, we will share the full dataset.
\textbf{Descriptions of real humans}
\begin{itemize}
\item A person with vitiligo, displaying unique patches of pigmentation on their face, with a confident, empowered presence.
\item A person with albinism, with pale skin, light hair, and striking, vibrant eyes.
\item An individual with facial scars or birthmarks, showing strength and resilience in their confident demeanor.
\item A person with a facial piercing, such as a septum ring, eyebrow piercing, or lip stud, adding an edgy, unconventional touch to their appearance.
\item A person with a full, bushy beard, groomed with care and reflecting a trendy, hipster style.
\end{itemize}
\textbf{Descriptions of fictional entities}
\begin{itemize}
\item Wicked witch with green skin, a crooked nose, and warts.
\item Queen of hearts with red heart-shaped lips and black heart under the eye.
\item Cyborg with metallic features and glowing eyes.
\item Haunted doll with cracked porcelain skin and blacked-out eyes.
\item Fairy tale princess with blond hair and rosy cheeks.
\end{itemize}
\textbf{Makeup and Dress-up styles}
\begin{itemize}
\item A mystical unicorn with pastel rainbow hair, a glitter horn, and sparkling pink and purple makeup.
\item An otherworldly alien with silver body paint, holographic eyes, and a metallic lip.
\item A dark sorceress with black lipstick, smoky purple eyeshadow, and a jeweled headpiece.
\item A steampunk robot with metallic gears, goggles, and metallic silver or bronze makeup.
\item A colorful peacock with blue and green feathers, bold eyeshadow, and bright pink lips.
\end{itemize}
All results shown in the paper are on avatars stylized using novel styles that were not used in the training dataset. In particular, for the test styles used for Figure 7 in the main paper, we use 25 novel art styles listed below, which the model did not see during training.
\begin{itemize}
\item Impressionist portrait painting
\item Black and white pen and ink portrait drawing
\item Lino portrait print
\item Realistic line art portrait drawing
\item Medieval portrait painting
\item Bronze statue
\item 3D render
\item Vintage anime chibi character
\item Graffiti portrait
\item Degas chalk portrait painting
\item Pointillism portrait drawing
\item Character design by hergé 
\item Cartoon
\item Oil painting
\item Pencil portrait drawing
\item Gouache portrait painting
\item Acrylic portrait painting
\item Watercolor painting
\item Pop art poster
\item Fauvism portrait painting
\item Mosaic
\item Caricature
\item 8-bit pixel art
\item Miro portrait painting
\item Abstract white cubist sculpture
\end{itemize}

\end{document}